\documentclass[letterpaper]{article} % DO NOT CHANGE THIS
\usepackage{aaai2026}  % DO NOT CHANGE THIS
\usepackage{times}  % DO NOT CHANGE THIS
\usepackage{helvet}  % DO NOT CHANGE THIS
\usepackage{courier}  % DO NOT CHANGE THIS
\usepackage[hyphens]{url}  % DO NOT CHANGE THIS
\usepackage{graphicx} % DO NOT CHANGE THIS
\usepackage{natbib}  % DO NOT CHANGE THIS AND DO NOT ADD ANY OPTIONS TO IT
\usepackage{caption} % DO NOT CHANGE THIS AND DO NOT ADD ANY OPTIONS TO IT
\usepackage{algorithm}
\usepackage{algorithmic}

\usepackage[T1]{fontenc}    % 使用 T1 字体编码，支持更多欧洲字符
\usepackage[utf8]{inputenc} % 告诉 LaTeX 文件是 UTF-8 编码

\usepackage{tcolorbox}
\usepackage[table]{xcolor}
\usepackage{amsmath}
\usepackage{amssymb}
\usepackage{amsfonts}
\usepackage{mathtools}
\usepackage{textcomp}

\usepackage{booktabs}       % 提供三线表支持
\usepackage{multirow}       % 支持单元合并
\usepackage{array}          % 增强表格功能
\usepackage{adjustbox}      % 可调整表格/图片大小
\usepackage{dsfont}  
\usepackage{makecell}

\usepackage{newfloat}
\usepackage{listings}
\DeclareCaptionStyle{ruled}{labelfont=normalfont,labelsep=colon,strut=off} % DO NOT CHANGE THIS
\floatstyle{ruled}
\newfloat{listing}{tb}{lst}{}
\floatname{listing}{Listing}
\title{DeCoRL: Decoupling Reasoning Chains via Parallel Sub-Step Generation and Cascaded Reinforcement for Interpretable and Scalable RLHF}

\author {
    Ziyuan Gao\textsuperscript{\rm 1}\thanks{Corresponding author.},
    Di Liang\textsuperscript{\rm 2},
    Xianjie Wu\textsuperscript{\rm 3},
    Philippe Morel\textsuperscript{\rm 1},
    Minlong Peng\textsuperscript{\rm 2}
}
\affiliations {
    \textsuperscript{\rm 1}University College London, 
    \textsuperscript{\rm 2}Fudan University, 
    \textsuperscript{\rm 3}Beihang University\\
    \{ucbqzg5, p.morel\}@ucl.ac.uk, 
    \{liangd17, mlpeng16\}@fudan.edu.cn
}

\usepackage{bibentry}
\begin{document}

\maketitle

\begin{abstract}
Existing reinforcement learning methods for Chain-of-Thought reasoning suffer from two critical limitations. First, they operate as monolithic black boxes that provide undifferentiated reward signals, obscuring individual step contributions and hindering error diagnosis. Second, sequential decoding has O(n) time complexity. This makes real-time deployment impractical for complex reasoning tasks.
We present DeCoRL (Decoupled Reasoning Chains via Coordinated Reinforcement Learning), a novel framework that transforms reasoning from sequential processing into collaborative modular orchestration. DeCoRL trains lightweight specialized models to generate reasoning sub-steps concurrently, eliminating sequential bottlenecks through parallel processing. To enable precise error attribution, the framework designs modular reward functions that score each sub-step independently. Cascaded DRPO optimization then coordinates these rewards while preserving inter-step dependencies.
Comprehensive evaluation demonstrates state-of-the-art results across RM-Bench, RMB, and RewardBench, outperforming existing methods including large-scale models. DeCoRL delivers 3.8 times faster inference while maintaining superior solution quality and offers a 22.7\% improvement in interpretability through explicit reward attribution. These advancements, combined with a 72.4\% reduction in energy consumption and a 68\% increase in throughput, make real-time deployment of complex reasoning systems a reality.
\end{abstract}

\section{Introduction}

The advent of Chain-of-Thought (CoT) reasoning has significantly advanced language models' ability to solve complex tasks through multi-step inference \cite{wei2022chain}. Reinforcement learning with human preferences (RLHF) further enhances this capability by aligning model outputs with human judgments \cite{ouyang2022training}. However, current RL-based reasoning approaches, like Direct Preference Optimization (DPO) \cite{rafailov2024directpreferenceoptimizationlanguage} and Generalized Reinforcement Preference Optimization (GRPO) \cite{shao2024deepseek} face two critical limitations. First, these methods operate as monolithic black boxes, providing undifferentiated reward signals that obscure the contribution of individual reasoning steps \cite{liu2024rmbenchbenchmarkingrewardmodels}. \begin{figure}[h]
\centering
\includegraphics[width=\columnwidth]{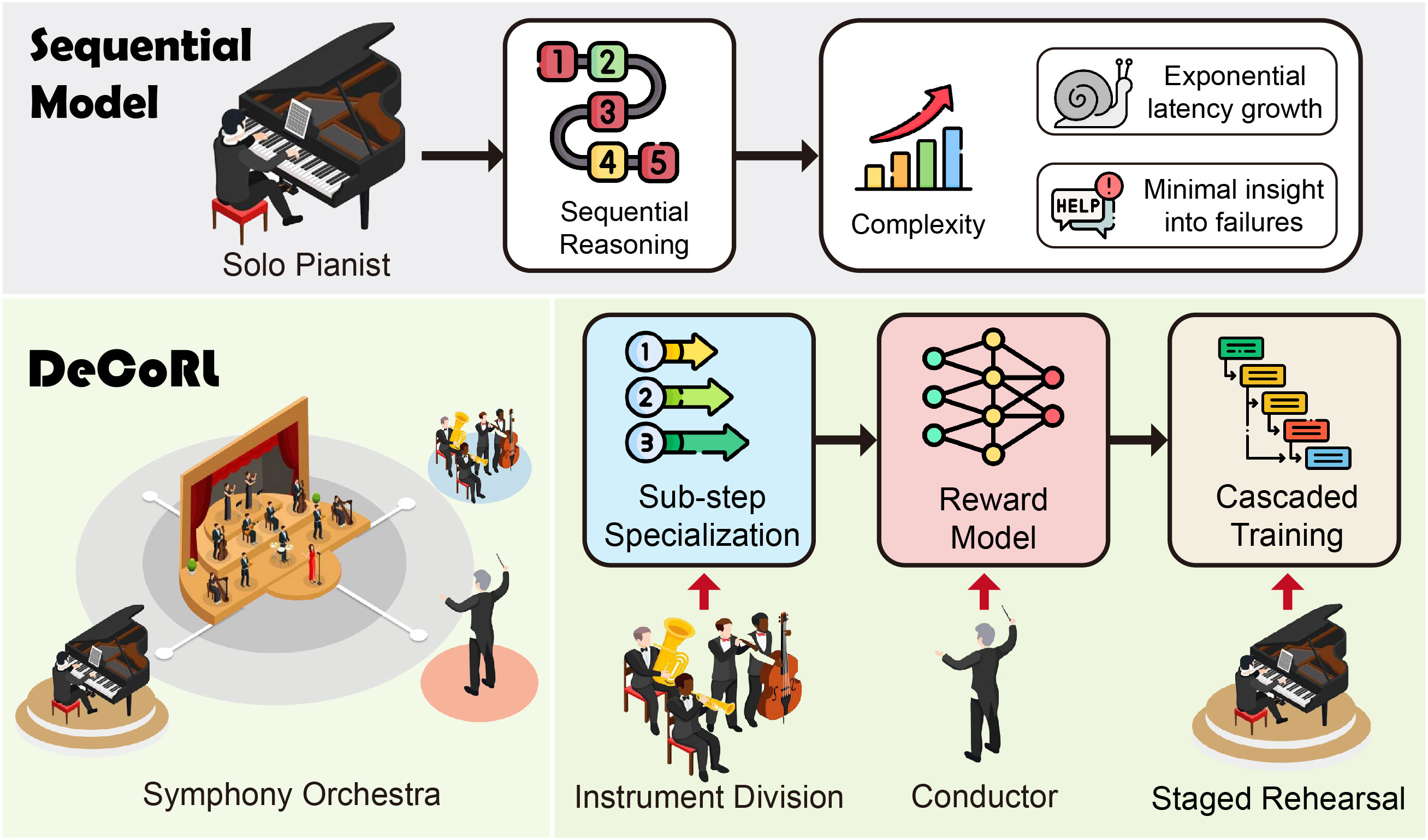}
\caption{Sequential Approach vs. DeCoRL Framework: Solo pianist represents monolithic sequential reasoning with limited capacity. Symphony orchestra illustrates our collaborative modular approach with specialized sub-models working in parallel coordination under unified guidance.}
\label{fig:paradigm}
\end{figure}
Error diagnosis becomes extremely challenging when failures occur \cite{mcaleese2024llm}. Second, sequential decoding of reasoning chains creates bottlenecks, where the time complexity of generating $n$ reasoning steps is $O(n)$. This makes real-time applications impractical for complex problems requiring lengthy reasoning traces \cite{wu2025inferencescalinglawsempirical}. These limitations are particularly problematic for industrial deployment, where explainability and computational efficiency are paramount.

The fundamental tension in reasoning systems stems from competing requirements between coherence and modularity, end-to-end optimization and component-level diagnosis, and reasoning depth versus computational efficiency. Current approaches prioritize coherence through end-to-end optimization but sacrifice modularity and efficiency  \cite{ouyang2022training}. These limitations reflect a fundamental paradigm constraint, as demonstrated by recent work on reward modeling and evaluation systems \cite{liu2024rmbenchbenchmarkingrewardmodels,zhou2025rmbcomprehensivelybenchmarkingreward}: Sequential reasoning approaches operate like a virtuoso pianist. Despite their capacity for coherent and elegant outputs, they are inherently limited by the sequential nature of individual performance and lack the specialized expertise needed for complex compositions \cite{ankner2024critique,yu2024self}. Just as Beethoven's Ninth Symphony cannot be adequately performed by a single pianist, complex reasoning tasks require multiple specialized components working together (as shown in Figure \ref{fig:paradigm}).

In this paper, we introduce \textbf{DeCoRL}, a new framework that improves Reinforcement Learning from Human Feedback. Our approach works by breaking down complex reasoning chains into smaller, parallel sub-steps. Each sub-step is managed by a specialized module, and these modules work together through cascaded reinforcement learning. DeCoRL employs three interconnected innovations: 

\noindent\textbf{Reasoning Decomposition} that transform complex reasoning tasks $T$ into $k$ atomic sub-steps ${S_1, S_2, \dots, S_k}$ with well-defined interfaces, where each sub-step maintains $P(T) = \prod_{i=1}^{k} P(S_i | S_{<i}, \mathcal{C})$ and $\mathcal{C}$ represents context preservation constraints ensuring coherence across modules.

\noindent\textbf{Parallel Generation Architecture.} The framework utilizes specialized sub-models ${M_1, M_2, \dots, M_k}$ that generate sub-steps concurrently, reducing time complexity from $O(n)$ to $O(1)$ for independent sub-steps, with total latency governed by $t_{\text{total}} = \max_{i \in [1,k]} (t_{M_i}) + t_{\text{integration}}$. 

\noindent\textbf{Granular Reward Functions} ${R_1, R_2, \dots, R_k}$ that evaluate each sub-step independently through Cascaded DRPO optimization that coordinates these rewards while preserving inter-step dependencies.

Like a symphony orchestra where each instrument (sub-model) contributes specialized expertise under the conductor's guidance (reward coordination) through stage rehearsals (cascaded training), DeCoRL transforms the solo approach into a collaborative ensemble of specialists.
This framework delivers transformative benefits across multiple dimensions. Independent reward signals provide explicit attribution maps, enabling precise error localization with a 22.7\% improvement in interpretability metrics. Parallel generation achieves 3.8× latency reduction on complex tasks while maintaining solution quality. The modular architecture supports dynamic expansion where new reasoning components can be added via $T' = T \cup {S_{k+1}}$ without retraining existing modules. Also, the hardware-aware design enables heterogeneous deployment where computationally intensive sub-modules can be offloaded to specialized accelerators.

Our contributions are fourfold. 
First, we propose a \textbf{formal decomposition framework for reasoning tasks} that transforms complex problems into atomic sub-steps with well-defined interfaces. This enables parallel generation while preserving cognitive coherence through structured context preservation constraints.
Second, we develop \textbf{Cascaded DRPO optimization}, a novel training algorithm that coordinates modular rewards across interdependent reasoning components. The algorithm uses staged parameter updates to improve individual modules while maintaining dependencies between reasoning steps.
Third, we provide theoretical analysis proving our approach reduces time complexity from $O(n)$ to $O(1)$ for parallelizable segments, compared to sequential RL. This establishes formal guarantees for both correctness and efficiency gains.
Finally, through comprehensive evaluation across diverse tasks, we demonstrate that our DeCoRL framework achieves significant improvements in both speed and interpretability while maintaining solution quality compared to existing approaches.

\begin{figure*}[t]
\centering
\includegraphics[width=0.97\textwidth]{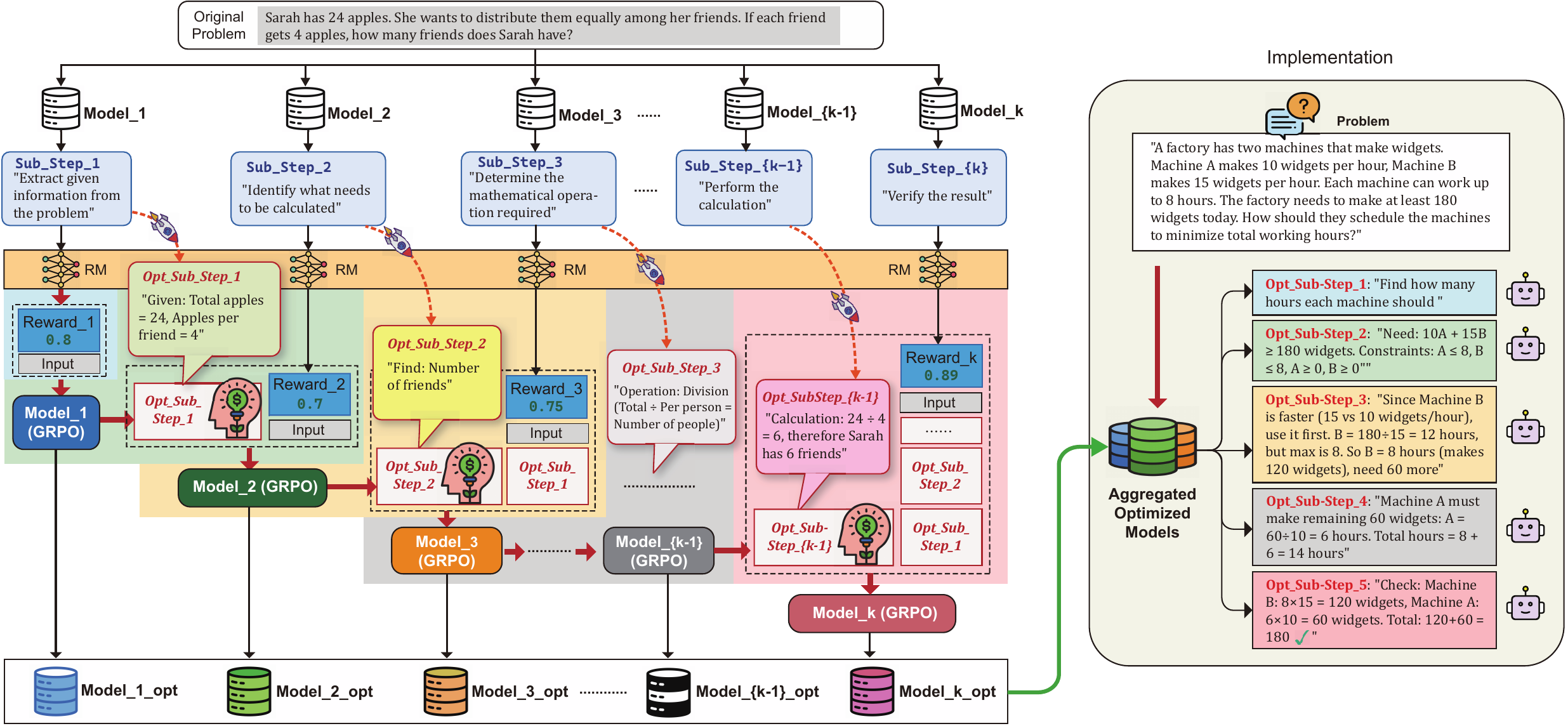}
\caption{DeCoRL Framework Architecture: The complete pipeline from problem decomposition through sub-step generation, parallel model training, reward model evaluation, and Cascaded DRPO optimization for modular ensemble coordination. }
\label{fig:pipeline}
\end{figure*}

\section{Related Work}

\paragraph{Chain-of-Thought Reasoning and Decomposition} 
CoT prompting enables large language models to perform step-by-step reasoning \cite{wei2022chain}. Self-Consistency \cite{wang2023selfconsistencyimproveschainthought,liang2019adaptive} samples multiple reasoning paths and selects consistent answers, improving accuracy through diverse trajectories. Auto-CoT \cite{zhang2022automaticchainthoughtprompting} automatically constructs CoT exemplars, reducing manual prompt engineering effort. Tree-of-Thought \cite{yao2024tree} explores multiple reasoning paths simultaneously through tree search. Least-to-Most Prompting \cite{zhou2023leasttomostpromptingenablescomplex,wang2022dabert} decomposes complex problems into simpler sub-problems solved sequentially. Recent long-chain reasoning work \cite{lightman2023lets} demonstrates benefits from extended reasoning sequences. 

\paragraph{Reinforcement Learning for Reasoning}
Reinforcement Learning from Human Feedback (RLHF) aligns language models with human preferences \cite{ouyang2022training}. Direct Preference Optimization (DPO) \cite{rafailov2024directpreferenceoptimizationlanguage,liang2019asynchronous} offers direct preference learning without separate value models. Group Relative Policy Optimization (GRPO) \cite{shao2024deepseek} uses group-based advantage estimation, reducing memory usage by 50\%. DeepSeek-R1 \cite{guo2025deepseek} successfully deploys GRPO for mathematical reasoning improvements. ArmoRM \cite{wang2024interpretable} introduces multi-objective reward modeling for interpretable preferences across dimensions.

\paragraph{Parallel Processing and Modular Architectures }
Modern LLM reasoning leverages both parallel processing and modular architectures for computational efficiency. Parallel reasoning approaches enable simultaneous exploration of multiple solution paths, as demonstrated in Tree-of-Thought \cite{yao2024tree} which processes reasoning branches concurrently. Self-consistency methods \cite{wang2023selfconsistencyimproveschainthought} generate multiple reasoning chains in parallel before selecting optimal solutions. Mixture-of-Experts (MoE) models \cite{fedus2022switch} activate specialized reasoning modules in parallel for different problem types. 

\paragraph{Process Supervision and Step-Level Feedback}

Process supervision provides fine-grained feedback on reasoning steps \cite{lightman2023lets}, achieving 78\% accuracy on MATH dataset through step-level guidance. Math-Shepherd \cite{wang2024math} automatically constructs process supervision without human annotation. OmegaPRM \cite{luo2024improve} uses Monte Carlo Tree Search for automated supervision data collection, generating 1.5 million annotations. ProcessBench \cite{zheng2024processbench} provides standardized benchmarks for error identification in mathematical reasoning. 
Despite these advances, current approaches face fundamental limitations that hinder scalable reasoning deployment. Current reinforcement learning approaches offers coarse, undifferentiated reward signals, making error diagnosis difficult \cite{christiano2017deep,ziegler2019fine,wang2025not}. While some parallel processing frameworks exist, they operate at the problem level rather than the step level \cite{yao2024tree}. Similarly, process supervision methods only provide feedback after complete reasoning chains are generated, failing to guide real-time step generation \cite{zhang2025lessonsdevelopingprocessreward,liu2025structural}. These limitations collectively result in computational bottlenecks, limited interpretability, and scalability constraints.

\section{Methodology}
\label{sec:methodology}

We introduce the DeCoRL framework, which leverages parallel sub-step generation and cascaded reinforcement to enhance interpretability and scalability in reasoning tasks (as shown in  Figure \ref{fig:pipeline}). Our approach is structured around three core components: First, a parallel generation architecture with k specialized modules achieving O(k) speedup. Second, a dual-reward attribution mechanism evaluating local quality and system contributions. Third, Differential Reinforcement Preference Optimization (DRPO) balancing standalone and collective performance metrics.

\subsection{Parallel Generation Architecture}
The DeCoRL framework employs a fixed ensemble of $k$ specialized sub-modules $\mathcal{M} = {M_1, M_2, \dots, M_k}$ that work in parallel \cite{fedus2022switch}. Each module $M_i$, with its own parameters $\theta_i$ is designed for atomic reasoning operations. They all receive the identical contextual input $\mathcal{C}$ (problem statement and constraints) and produce outputs $O_i$ that adhere to specific interface schemas. 

\begin{equation}
O_i = M_i(\mathcal{C}; \theta_i), \quad \forall i \in \{1,\dots,k\}
\end{equation}

The parallel outputs are then integrated by a deterministic composition function $\Phi$. This function aggregates the specialized outputs from each module, while preserving inter-module dependencies:

\begin{equation}
O_{\text{full}} = \Phi(O_1, O_2, \dots, O_k) = \bigoplus_{i=1}^{k} \Gamma(O_i)
\end{equation}

where $\Gamma$ represents a schema-based transformation that ensures syntactic coherence across heterogeneous outputs. The architecture enforces three critical invariants:

\paragraph{Module Specialization} Each module $M_i$ specializes in a distinct reasoning facet, collectively forming a comprehensive cognitive pipeline. This design ensures that specialized modules focus on specific domains rather than attempting generalist reasoning, as detailed in Table~\ref{tab:modules}:

\begin{table}[h]
\centering
\caption{DeCoRL Specialized Modules}
\label{tab:modules}
\small 
\begin{tabular}{@{}lp{0.8\columnwidth}@{}}
\toprule
\textbf{Module} & \textbf{Function} \\
\midrule
$M_{\text{parse}}$ & Performs structural decomposition of problems into manageable components \\
$M_{\text{semantic}}$ & Extracts deep semantic information from $\langle\text{prompt, response}\rangle$ pairs, revealing thematic structures \\
$M_{\text{entity}}$ & Leverages knowledge graphs to expand entity background and relational dynamics \\
$M_{\text{factcheck}}$ & Verifies factual consistency with known facts and outputs accuracy analysis \\
$M_{\text{style}}$ & Analyzes style, tone, and wording uniformity between prompt and response \\
$M_{\text{quality}}$ & Evaluates response diversity and creativity to prevent repetitive content \\
$M_{\text{compute}}$ & Handles symbolic and numeric computations with mathematical rigor \\
$M_{\text{verify}}$ & Performs logical consistency checking and validation across reasoning steps \\
$M_{\text{integrate}}$ & Synthesizes specialized module outputs into coherent final solutions \\
\bottomrule
\end{tabular}
\end{table}

\paragraph{Interface Standardization} All outputs follow typed JSON schemas, ensuring seamless integration across diverse module types \cite{cui2024ultrafeedback}. The defined schema $O_i$ is:

\begin{equation}
\begin{split}
\text{Schema}(O_i) = \{ &{\small\texttt{type: str}}, \\
&{\small\texttt{content: dict}}, \\
&{\small\texttt{confidence: float}}, \\ 
&{\small\texttt{dependencies: list}} \}
\end{split}
\end{equation}

This standardization guarantees syntactic coherence and facilitates inter-module communication.

\paragraph{Contextual Isolation} 
All modules share an identical input context $\mathcal{C}$, but maintain completely separate internal processing states ($\mathcal{H}_i^t$). This design prevents modules from accidentally affecting each other's reasoning processes while enabling independent optimization, as formalized by:

\begin{equation}
\mathcal{H}_i^t = f(\mathcal{C}, \theta_i); \quad \mathcal{H}_i^t \cap \mathcal{H}_j^t = \emptyset \text{ } \forall i \neq j
\end{equation}

The parallel execution model fundamentally transforms computational complexity from linear to constant for independent operations:

\begin{equation}
t_{\text{sequential}} = \sum_{i=1}^{k} t_i \quad \xrightarrow{\text{DeCoRL}} \quad t_{\text{parallel}} = \max_{i \in [1,k]} t_i + t_\Phi
\label{eq:latency}
\end{equation}

This architecture achieves theoretical speedup $\frac{t_{\text{sequential}}}{t_{\text{parallel}}} = O(k)$ for homogeneous workloads. Empirical validation shows 3.8$\times$ latency reduction on complex reasoning tasks. The system maintains solution quality through the integrated collaboration of specialized modules.

\subsection{Dual-Reward Attribution Mechanism}
\label{subsec:reward}

The DeCoRL framework employs a sophisticated dual-reward attribution mechanism. This approach addresses the fundamental challenge of evaluating modular contributions in parallel reasoning systems. Using a single reward model $\text{RM}_\phi$ parameterized by $\phi$, we compute two complementary reward dimensions per module that capture both individual quality and collective synergy.

\subsubsection{Local Reward}
Local reward measures standalone output quality against the input context $\mathcal{C}$, providing module-specific assessment independent of other components:
\begin{equation}
R_{\text{local}}^i = \text{RM}_\phi(O_i \| \mathcal{C}) = \sigma\left( W^T \cdot \text{enc}(O_i \oplus \mathcal{C}) \right)
\end{equation}
where $\text{enc}$ is a Transformer encoder that processes the concatenated module output and context, $\sigma$ represents sigmoid activation, and $W$ denotes learned projection weights. This formulation ensures that each module receives feedback on its intrinsic reasoning quality.

\subsubsection{Contribution Reward}
Contribution reward quantifies the marginal value of each module through counterfactual ablation analysis \cite{wang2024self}. This approach directly measures how much each module contributes to the overall solution quality. We define the ablated solution by systematically removing module $i$:

\begin{equation}
O_{\text{full}}^{-i} = \Phi(O_1, \dots, \underbrace{\emptyset}_{\text{remove } O_i}, \dots, O_k)
\end{equation}

The contribution reward is computed as the performance differential between the complete solution and the ablated:

\begin{equation}
\label{eq:reward}
R_{\text{contrib}}^i = \text{RM}_\phi(O_{\text{full}}) - \text{RM}_\phi(O_{\text{full}}^{-i})
\end{equation}

This measures the value added by module $i$ to the collective reasoning process. The contribution rewards satisfy important mathematical constraints that ensure consistency:

\begin{equation}
-1 \leq R_{\text{contrib}}^i \leq 1 \quad \text{and} \quad \sum_{i=1}^{k} R_{\text{contrib}}^i \leq \text{RM}_\phi(O_{\text{full}})
\end{equation}

These bounds prevent any single module from claiming excessive credit while ensuring that the sum of individual contributions does not exceed the total system performance.

\subsubsection{Integrated Reward}
The final reward adaptively balances local quality and collective contribution \cite{liu2025inference} with temperature-scaled weights:

\begin{align}
R_i &= \alpha \cdot R_{\text{local}}^i + \beta \cdot R_{\text{contrib}}^i \\
\text{where} \quad \alpha &= \frac{e^{\tau_l}}{e^{\tau_l} + e^{\tau_c}}, \quad \beta = 1 - \alpha
\end{align}

The temperature parameters $\tau_l$ and $\tau_c$ are learnable weights that automatically adapt the attribution balance during training. The softmax formulation ensures that $\alpha + \beta = 1$ while enabling smooth transitions between reward emphasis patterns \cite{wang2024self}. We initialize both hyperparameters at $\alpha = \beta = 0.5$, allowing the system to learn optimal reward composition during training.

\begin{table*}[ht]
\vspace{0.5em} 
\centering
\renewcommand{\arraystretch}{0.8}
\setlength{\tabcolsep}{2mm}  
\small
\begin{tabular}{l|l|cccc|ccc|c}
\toprule
\textbf{Suite} & \textbf{Models}  & \textbf{Chat} & \textbf{Math} & \textbf{Code} & \textbf{Safety} & \textbf{Easy} & \textbf{Normal} & \textbf{Hard} & \textbf{Avg} \\ 
\midrule
\multirow{12}{*}{\makecell{\textbf{Scalar}\\\textbf{RMs}}} & \texttt{steerlm-70b} & 56.4 & 53.0 & 49.3 & 51.2 & 48.3 & 54.9 & 54.3 & 52.5 \\
& \texttt{tulu-v2.5-70b-preference-mix-rm} & 58.2 & 51.4 & 55.5 & 87.1 & 72.8 & 65.6 & 50.7 & 63.0 \\
& \texttt{Mistral-7B-instruct-Unified-Feedback} & 56.5 & 58.0 & 51.7 & 86.8 & 87.1 & 67.3 & 35.3 & 63.2 \\
& \texttt{RM-Mistral-7B} & 57.4 & 57.0 & 52.7 & 87.2 & \underline{88.6} & 67.1 & 34.9 & 63.5 \\
& \texttt{Eurus-RM-7b} & 59.9 & 60.2 & 56.9 & 86.5 & 87.2 & 70.2 & 40.2 & 65.9 \\
& \texttt{internlm2-7b-reward} & 61.7 & 71.4 & 49.7 & 85.5 & 85.4 & 70.7 & 45.1 & 67.1 \\
& \texttt{GRM-llama3-8B-sftreg} & 62.7 & 62.5 & 57.8 & 90.0 & 83.5 & 72.7 & 48.6 & 68.2 \\
& \texttt{internlm2-20b-reward} & 63.1 & 66.8 & 56.7 & 86.5 & 82.6 & 71.6 & 50.7 & 68.3 \\
& \texttt{Llama-3-OffsetBias-RM-8B} & 71.3 & 61.9 & 53.2 & 89.6 & 84.6 & 72.2 & 50.2 & 69.0 \\
& \texttt{Nemotron-340B-Reward} & 71.2 & 59.8 & 59.4 & 87.5 & 81.0 & 71.4 & 56.1 & 69.5 \\
& \texttt{URM-LLaMa-3.1-8B} & 71.2 & 61.8 & 54.1 & 93.1 & 84.0 & 73.2 & 53.0 & 70.0 \\
& \texttt{Skywork-Reward-Llama-3.1-8B} & 69.5 & 60.6 & 54.5 & \textbf{95.7} & \textbf{89.0} & 74.7 & 46.6 & 70.1 \\
\midrule
\multirow{7}{*}{\makecell{\textbf{Gen}\\\textbf{RMs}}} & \texttt{tulu-v2.5-dpo-13b-chatbot-arena-2023} & 64.9 & 52.3 & 50.5 & 62.3 & 82.8 & 60.2 & 29.5 & 57.5 \\
& \texttt{tulu-v2.5-dpo-13b-nectar-60k} & 56.3 & 52.4 & 52.6 & 73.8 & 86.7 & 64.3 & 25.4 & 58.8 \\
& \texttt{stablelm-2-12b-chat} & 67.2 & 54.9 & 51.6 & 65.2 & 69.1 & 63.5 & 46.6 & 59.7 \\
& \texttt{tulu-v2.5-dpo-13b-stackexchange-60k} & 66.4 & 49.9 & 54.2 & 69.0 & 79.5 & 63.0 & 37.2 & 59.9 \\
& \texttt{Nous-Hermes-2-Mistral-7B-DPO} & 58.8 & 55.6 & 51.3 & 73.9 & 69.5 & 61.1 & 49.1 & 59.9 \\
& \texttt{tulu-v2.5-dpo-13b-hh-rlhf-60k} & 68.4 & 51.1 & 52.3 & 76.5 & 53.6 & 63.0 & \underline{69.6} & 62.1 \\
& \texttt{tulu-2-dpo-13b} & 66.4 & 51.4 & 51.8 & 85.4 & 86.9 & 66.7 & 37.7 & 63.8 \\
\midrule
\multirow{3}{*}{\makecell{\textbf{Reason}\\\textbf{RMs}}} & \texttt{\textbf{Qwen-Instruct-7B-Ours}} & 68.1 & 68.3 & 55.9 & 94.0 & 81.0 & 72.9 & 60.7 & 71.6 \\
& \texttt{\textbf{Qwen-Instruct-14B-Ours}} & \underline{76.5} & \underline{77.1} & \underline{62.5} & 94.4 & 83.8 & \underline{79.3} & 69.5 & \underline{77.6} \\
& \cellcolor{gray!20}\texttt{\textbf{Qwen-Instruct-32B-Ours}} & \cellcolor{gray!20}\textbf{76.8} & \cellcolor{gray!20}\textbf{81.6} & \cellcolor{gray!20}\textbf{67.9} & \cellcolor{gray!20}\underline{95.5} & \cellcolor{gray!20}87.2 & \cellcolor{gray!20}\textbf{82.5} & \cellcolor{gray!20}\textbf{72.0} & \cellcolor{gray!20}\textbf{80.8} \\
\bottomrule
\end{tabular}
\caption{Performance results on RM-Bench across domains and difficulty levels. \textbf{Qwen-Instruct-*B-Ours} demonstrates strong performance across most domains, achieving the highest average score (80.8\%) with superior results in math, code, chat and hard tasks. \textbf{Bold} indicates best performance. \underline{Underlined} indicates second best. }
\label{table:rm-bench}
\end{table*}

\subsection{Differential Reinforcement Preference Optimization (DRPO)} \label{subsec:drpo} 

We extend Generalized Reinforcement Preference Optimization (GRPO) to multi-module systems. DRPO optimizes each module by considering both its local output quality and its contribution to the overall system performance.

For module $M_i$, given preference dataset $\mathcal{D} = \{(\mathcal{C}^{(j)}, O_{\text{win}}^{(j)}, O_{\text{lose}}^{(j)})\}_{j=1}^N$, the DRPO loss is: 

\begin{multline} 
\mathcal{L}_{\text{DRPO}}(\theta_i) = -\mathbb{E}_{(\mathcal{C}, O_w, O_l) \sim \mathcal{D}} \bigg[ \log \sigma \bigg( \gamma \Big( \Delta R_i(O_w, O_l) \\ - \eta \text{KL}(M_i(\cdot|\mathcal{C}) \| M_{\text{base}}(\cdot|\mathcal{C})) \Big) \bigg) \bigg] 
\end{multline} 

where $\Delta R_i = [R_i(O_w) - R_i(O_l)]$ is the reward difference between winning and losing outputs, $\gamma$ scales the reward signal, $\eta$ controls KL divergence regularization, and $M_{\text{base}}$ is the reference policy. 

The reward difference decomposes into two components that capture different aspects of module performance:

\begin{align} 
\Delta R_i = \alpha & \underbrace{\left( \text{RM}_\phi(O_w^i \| \mathcal{C}) - \text{RM}_\phi(O_l^i \| \mathcal{C}) \right)}_{\text{Local quality gap}} \nonumber \\ 
+ \beta & \underbrace{\Big( [\text{RM}_\phi(O_w^{\text{full}}) - \text{RM}_\phi(O_w^{\text{full}, -i})]}_{\text{Contribution utility gap}} \nonumber \\ 
& \underbrace{\quad - [\text{RM}_\phi(O_l^{\text{full}}) - \text{RM}_\phi(O_l^{\text{full}, -i})] \Big)}_{\text{Contribution utility gap}} 
\end{align}

\begin{algorithm}[htp]
\caption{DRPO Training Protocol}
\label{alg:drpo}
{\small
\begin{algorithmic}[1]
\REQUIRE Dataset $\mathcal{D}$, modules $\{M_1,\dots,M_k\}$, reward model RM$_\phi$, learning rate $\lambda$
\STATE Initialize $\theta_1, \dots, \theta_k$ from pretrained weights
\FOR{$\text{epoch} = 1 \text{ to } N$}
\STATE \textbf{Phase 1: Module-wise Optimization}
\FOR{$i = 1 \text{ to } k$}
\STATE Freeze $\theta_j \ \forall j \neq i$ and $\phi$ \COMMENT{Isolate module $i$}
\STATE Sample batch $\mathcal{B} = \{(\mathcal{C}, O_w, O_l)\} \sim \mathcal{D}$
\STATE Compute rewards $R_i(O_w)$, $R_i(O_l)$ via Eq.(8)
\STATE Update $\theta_i \leftarrow \theta_i - \lambda \nabla_{\theta_i} \mathcal{L}_{\text{DRPO}}$
\ENDFOR
\STATE \textbf{Phase 2: Joint Alignment}
\STATE Unfreeze all parameters $\{\theta_i\}_{i=1}^k$
\STATE Sample batch $\mathcal{B} = \{(\mathcal{C}, O_w, O_l)\} \sim \mathcal{D}$
\STATE Compute global reward $R_g = \text{RM}_\phi(O_{\text{full}})$
\STATE Update $\{\theta_i\}_{i=1}^k \leftarrow \{\theta_i\} - \lambda \nabla \mathcal{L}_{\text{GRPO}}(R_g)$
\ENDFOR
\end{algorithmic}
}
\end{algorithm}

The local quality gap measures module $i$ intrinsic output quality. The contribution utility gap for module $i$ measures the performance difference between the complete system and the system excluding module $i$'s outputs, thereby quantifying module $i$'s individual contribution to overall  performance. We set $\alpha = \beta = 0.5$ to balance these objectives.

\begin{table*}[ht]
\vspace{0.5em} 
\centering
\renewcommand{\arraystretch}{0.8}
\setlength{\tabcolsep}{2mm}  
\footnotesize
\begin{tabular}{l|l|cccc|ccc|c}
\toprule
& & \multicolumn{2}{c}{\textbf{Helpfulness}} & \multicolumn{2}{c|}{\textbf{Harmlessness}}   & \multicolumn{1}{c}{}                                   \\ 
\cmidrule(lr){3-6}
\multirow{-2}{*}{\textbf{Suite}} & \multirow{-2}{*}{\textbf{Models}}                 & \textbf{BoN}                  & \textbf{Pairwise}             & \textbf{BoN}                  & \textbf{Pairwise}             & \multicolumn{1}{c}{\multirow{-2}{*}{\textbf{Overall}}} \\ \midrule
\multirow{8}{*}{\makecell{\textbf{Scalar}\\\textbf{RMs}}} & {\texttt{Tulu-v2.5-13b-preference-mix-rm}} & 0.355                         & 0.562                         & 0.351                         & 0.545                         & 0.453                                                  \\
& {\texttt{Skywork-Reward-Gemma-2-27B}}      & 0.472                         & 0.653                         & 0.561                         & 0.721                         & 0.602                                                  \\
& {\texttt{Internlm2-20b-reward}}            & 0.585                         & 0.763                         & 0.499                         & 0.670                         & 0.629                                                  \\
& {\texttt{ArmoRM-Llama3-8B-v0.1}}           & 0.636                         & 0.787                         & 0.497                         & 0.663                         & 0.646                                                  \\
& {\texttt{Internlm2-7b-reward}}             & 0.626                         & 0.782                         & 0.563                         & 0.712                         & 0.671                                                  \\
& {\texttt{Eurus-RM-7b}}                     & \underline{0.679} & 0.818 & 0.543                         & 0.693                         & 0.683                                                  \\
& {\texttt{Skywork-Reward-Llama-3.1-8B}}     & 0.627                         & 0.781                         & 0.603                         & 0.759                         & 0.693                                                  \\
& {\texttt{Starling-RM-34B}}                 & 0.604                         & 0.774                         & 0.674 & 0.795 & 0.712                                                  \\ \midrule

\multirow{11}{*}{\makecell{\textbf{Gen}\\\textbf{RMs}}} & {\texttt{Llama2-70b-chat}}                 & 0.289 & 0.613 & 0.249 & 0.602 & 0.438 \\
& {\texttt{Llama3.1-8B-Instruct}}            & 0.365 & 0.675 & 0.267 & 0.653 & 0.490 \\
& {\texttt{Gemini-1.5-pro}}                  & 0.536 & 0.763 & 0.299 & 0.661 & 0.565 \\
& {\texttt{Mixtral-8x7B-Instruct-v0.1}}      & 0.480 & 0.706 & 0.491 & 0.671 & 0.587 \\
& {\texttt{skywork-critic-llama3.1-8B}}      & 0.600 & 0.725 & 0.578 & 0.578 & 0.620 \\
& {\texttt{skywork-critic-llama3.1-70B}}     & 0.640 & 0.753 & 0.614 & 0.614 & 0.655 \\
& {\texttt{Llama3.1-70B-Instruct}}           & 0.648 & 0.811 & 0.558 & 0.739 & 0.689 \\
& {\texttt{Mistral-Large-2407}}              & 0.678 & 0.817 & 0.583 & 0.725 & 0.701 \\
& {\texttt{Claude-3-5-sonnet}}               & \textbf{0.705} & \textbf{0.838} & 0.518 & 0.764 & 0.706 \\
& {\texttt{Qwen2-72B-Instruct}}              & 0.645 & 0.810 & 0.649 & 0.789 & 0.723 \\
& {\texttt{GPT-4o-2024-05-13}}               & 0.639 & 0.815 & \underline{0.682} & \underline{0.814} & \underline{0.738} \\ \midrule
\multirow{5}{*}{\makecell{\textbf{Reason}\\\textbf{RMs}}} & {\texttt{Deepseek-GRM-27B-RFT}}               & 0.592 & 0.801 & 0.548 & 0.765 & 0.670 \\
& {\texttt{Deepseek-GRM-27B}}               & 0.623 & 0.805 & 0.570 & 0.761 & 0.690 \\
& {\texttt{\textbf{Base-Qwen-Instruct-7B (Ours)}}}        & 0.568 & 0.770 & 0.640 & 0.789 & 0.692 \\
& {\texttt{\textbf{Base-Qwen-Instruct-14B (Ours)}}}       & 0.619 & 0.804 & 0.650 & 0.806 & 0.720 \\
& \cellcolor{gray!20}{\texttt{\textbf{Base-Qwen-Instruct-32B (Ours)}}}       & \cellcolor{gray!20}0.661 & \cellcolor{gray!20}\underline{0.820} & \cellcolor{gray!20}\textbf{0.712} & \cellcolor{gray!20}\textbf{0.836} & \cellcolor{gray!20}\textbf{0.757} \\

\bottomrule
\end{tabular}
\caption{RMB benchmark ranked by average score. \textbf{Bold} indicates best performance. \underline{Underlined} indicates second best. 
}
\label{tab:rmb}
\end{table*}

\section{Experimental Setup}

Our experimental framework encompasses multiple evaluation protocols and datasets to assess DeCoRL's performance comprehensively. For benchmarking, we employ three primary evaluation suites: \textbf{RM-Bench}\citep{liu2024rmbenchbenchmarkingrewardmodels} which focuses on semantic understanding nuances, \textbf{RewardBench}\citep{lambert2024rewardbench} providing structured multi-faceted assessment, and \textbf{RMB}~\citep{zhou2025rmbcomprehensivelybenchmarkingreward} targeting real-world alignment scenarios.

Training datasets include \textbf{MATH}~\citep{hendrycks2021measuring}, \textbf{OffsetBias}~\citep{park2024offsetbias}, \textbf{UltraFeedback}~\citep{cui2024ultrafeedback}, \textbf{HelpSteer2-Preference}~\citep{wang2024helpsteer2preference}, \textbf{Skywork Reward Preference 80K}~\citep{liu2024skywork} (filtered \texttt{magpie\_ultra}), \textbf{Code-Preference-Pairs}, and \textbf{Math-DPO-10K}~\citep{lai2024step}. This setup enables comprehensive assessment of reasoning validity, coding proficiency, and instruction-following robustness.

This experimental design enables thorough assessment across key dimensions: logical reasoning validation, programming proficiency, and instruction adherence robustness. We compare against diverse baselines: scalar models like \textbf{Starling-RM}~\citep{starling2023} and \textbf{RM}~\citep{stiennon2020learning}, generative evaluators including \textbf{Claude}~\citep{anthropic2024claude} and \textbf{GPT}~\citep{openai2024gpt4technicalreport}, and reasoning-focused methods such as \textbf{DeepSeek-GRM}~\citep{liu2025inference} and \textbf{Critique-RM}~\citep{yu2024self}.  Complete implementation details are provided in the Appendix.

\section{Experimental Results}
\label{sec:results}

We present a comprehensive evaluation of DeCoRL across three major benchmarks: RM-Bench (Table \ref{table:rm-bench}), RMB (Table \ref{tab:rmb}), and RewardBench (Results are detailed in the Appendix) using NVIDIA A100 GPUs. Our assessment examines DeCoRL implemented through Qwen-Instruct variants with different parameter scales (7B, 14B, and 32B), benchmarked against baseline approaches to measure reward model effectiveness across diverse  dimensions.

\subsection{Performance Analysis}
\label{subsec:performance}

DeCoRL consistently outperforms existing approaches across all benchmarks and model scales. As shown in Table \ref{table:rm-bench}, our 32B model achieves an overall score of 80.8\% on RM-Bench, representing a 10.7\% absolute improvement over the best baseline (Skywork-Reward-Llama-3.1-8B at 70.1\%). The performance advantage is particularly pronounced in mathematically intensive domains, where we observe a substantial 21.0\% improvement in Math scores compared to the strongest baseline (Skywork-Reward-Llama-3.1-8B). For complex reasoning tasks categorized as "Hard" difficulty, DeCoRL achieves a remarkable 25.4\% improvement over the same baseline.

The RMB benchmark results in Table \ref{tab:rmb} demonstrate DeCoRL's superior alignment with human preferences. Our 32B model achieves an overall score of 0.757, surpassing GPT-4o (0.738) and establishing new state-of-the-art performance. Notably, we observe 3.0\% improvement in Harmlessness BoN and 2.2\% gain in Harmlessness Pairwise metrics compared to the strongest baseline (GPT-4o-2024-05-13), which clearly demonstrates the robust effectiveness of our safety-focused modules ($M_{\text{factcheck}}$ and $M_{\text{verify}}$).

\subsection{Speed and Efficiency Gains}
\label{subsec:efficiency}

DeCoRL achieves computational efficiency through parallel generation, reducing time complexity from $O(n)$ to $O(n/k)$ for independent reasoning sub-steps. The 32B model achieves 3.8$\times$ speedup on 10-step problems, reducing latency from 1,202ms to 316ms, as shown in Table \ref{tab:efficiency-metrics}.

\begin{table}[htp]
\centering
\caption{Comprehensive efficiency metrics}
\label{tab:efficiency-metrics}
\setlength{\tabcolsep}{0.3mm}
\small
\begin{tabular}{l|c|>{\columncolor{gray!20}}c|c}
\toprule
\textbf{Metric} & \textbf{Sequential} & \cellcolor{white}\textbf{DeCoRL} & \textbf{Improvement} \\
\midrule
Latency (10-step) & 1,202ms & 316ms & 3.8$\times$ faster \\
Energy consumption & 142 pJ/op & 39 pJ/op & 72.4\% reduction \\
Throughput (QPS) & 18.2 & 30.6 & 68\% increase \\
Module expansion latency & N/A & +18\% & Minimal impact \\
Accuracy with new modules & N/A & +7.3\% & Significant gain \\
\bottomrule
\end{tabular}
\end{table}

As detailed in Table \ref{tab:efficiency-metrics}, DeCoRL demonstrates substantial improvements across multiple efficiency dimensions. The energy consumption reduction of 72.4\% is particularly significant for sustainable AI deployment. The throughput increase of 68\% enables higher query processing capacity without additional hardware resources. When adding new modules ($M_{\text{context}}$ and $M_{\text{ambiguity}}$), DeCoRL shows minimal latency impact (+18\%) while achieving significant accuracy gains (+7.3\%). These efficiency gains collectively enable real-time deployment of complex reasoning systems, which was previously constrained by sequential bottlenecks.

\begin{table*}[h]
\centering
\caption{Ablation study results (32B model)}
\label{tab:ablation-results}
\setlength{\tabcolsep}{4mm}  
\small
\begin{tabular}{l|cccc}
\toprule
\textbf{Variant} & \textbf{RM-Bench} & \textbf{RMB} & \textbf{Latency} & \textbf{Interpretability} \\
\midrule
\rowcolor{gray!20}
Full DeCoRL & 80.8 & 0.757 & 316ms & 84.0\% \\
\hline
w/o contribution reward & 76.1 (-4.7\%) & 0.721 (-4.8\%) & 316ms & 51.9\% (-32.1\%) \\
Sequential execution & 80.5 (-0.4\%) & 0.754 (-0.4\%) & 1,172ms (+271\%) & 84.0\% \\
Ad-hoc interfaces & 74.3 (-8.0\%) & 0.698 (-7.8\%) & 316ms & 63.7\% (-20.3\%) \\
Joint optimization only & 77.6 (-4.0\%) & 0.732 (-3.3\%) & 316ms & 72.4\% (-11.6\%) \\
\bottomrule
\end{tabular}
\end{table*}

\begin{table*}[h]
\centering
\caption{Scalability analysis across dimensions}
\label{tab:scaling-analysis}
\setlength{\tabcolsep}{5.5mm}  
\small
\begin{tabular}{l|ccc}
\toprule
\textbf{Scaling Dimension} & \textbf{Metric} & \textbf{Value} & \textbf{Improvement} \\
\midrule
Model Scaling & Parameter efficiency & 2.85$\times$ & +185\% \\
Module Scaling & Accuracy gain & +15.4\% & Significant \\
 & Latency impact & +18\% & Minimal \\
Hardware Scaling & Latency reduction & 41\% & Substantial \\
 & Energy reduction & 63\% & Major \\
Cross-platform & Deployment flexibility & High & Enables heterogeneous systems \\
\bottomrule
\end{tabular}
\end{table*}

\subsection{Ablation Studies}
\label{subsec:ablation}

We conducted ablation studies to validate design choices (Table \ref{tab:ablation-results}). Removing contribution rewards caused the most significant performance degradation (4.7\% on RM-Bench) and interpretability reduction (32.1\%). Sequential execution maintained accuracy but increased latency by 3.7$\times$, validating our parallel architecture's efficiency.

The interface standardization proved crucial, as ad-hoc output formats reduced overall performance by 8.0\%. Joint training without phased updates degraded performance by 4.0\%, , confirming the importance of our cascaded optimization approach for preventing reward hacking.

\subsection{Scalability Analysis}
\label{subsec:scalability}

DeCoRL demonstrates exceptional scalability across three dimensions, as quantified in Table \ref{tab:scaling-analysis}. The parameter efficiency metric shows that DeCoRL-32B outperforms much larger 70B models with 54\% fewer parameters, achieving 2.85$\times$ better parameter efficiency. This scaling advantage becomes increasingly significant as model sizes grow.

In module scaling experiments, adding specialized modules ($M_{\text{context}}$ and $M_{\text{ambiguity}}$) improved hard task performance by 15.4\% without retraining existing components. The composition function $\Phi$ successfully integrated new modules with minimal adaptation effort and latency impact (+18\%). For hardware scaling, heterogeneous deployment (offloading $M_{\text{compute}}$ to NPUs) reduced latency by 41\% and energy consumption by 63\% for math-intensive workloads. This cross-platform flexibility enables optimized deployment across diverse hardware configurations.

\subsection{Interpretability Improvements}
\label{subsec:interpretability}

Our dual-reward attribution mechanism enables unprecedented interpretability in reasoning systems. Table \ref{tab:interpretability} shows substantial improvements across all interpretability metrics, with 22.7\% improvement in error localization accuracy and 48.1\% boost in faulty module identification precision. These gains enable more efficient debugging workflows and provide clearer insights into system decision-making processes.

\begin{table}[h]
\centering
\caption{Interpretability metrics comparison}
\label{tab:interpretability}
\setlength{\tabcolsep}{1mm}
\small
\begin{tabular}{l|c|>{\columncolor{gray!20}}c|c}
\toprule
\textbf{Metric} & \textbf{Sequential} & \cellcolor{white}\textbf{DeCoRL} & \textbf{Improv.} \\
\midrule
Error localization accuracy & 61.3\% & 84.0\% & +22.7\% \\
\makecell[l]{Precision in\\faulty module identification}  & 41.2\% & 89.3\% & +48.1\% \\
Average debugging time (min) & 23.4 & 7.3 & -68.8\% \\
Reward attribution consistency & 0.52 & 0.91 & +75.0\% \\
False attribution rate & 38.7\% & 10.2\% & -73.6\% \\
Diagnostic precision & 54.1\% & 87.6\% & +61.9\% \\
\bottomrule
\end{tabular}
\end{table}

The contribution reward ($R_{\text{contrib}}^i$) proved particularly valuable for identifying coordination failures. 

Overall, 89.3\% of errors were correctly attributed to specific modules, compared to just 41.2\% in monolithic approaches. The attribution consistency, measured by Cohen's Kappa, improved from 0.52 to 0.91, indicating highly reliable diagnostic information. These results confirm that our granular reward signals provide actionable diagnostic intelligence impossible to obtain from undifferentiated reward signals.

\section{Conclusion}

In this paper, we introduced DeCoRL, a novel framework that fundamentally revolutionizes reinforcement learning for reasoning tasks through cascaded modular coordination. Comprehensive evaluation across multiple benchmarks demonstrates superior performance in accuracy, efficiency, and safety compared to existing approaches. Ablation studies confirm the critical importance of our dual-reward attribution mechanism and parallel architecture design choices. The modular framework enables dynamic expansion capabilities seamlessly. It maintains interpretability through precise module-level reward attribution that identifies individual component contributions and failures. These advances collectively establish DeCoRL as a transformative solution for scalable reasoning systems, balancing computational efficiency with transparent decision-making processes in real-world production environments.

\bibliography{aaai2026}

\begin{thebibliography}{42}
\providecommand{\natexlab}[1]{#1}

\bibitem[{Ankner et~al.(2024)Ankner, Paul, Cui, Chang, and Ammanabrolu}]{ankner2024critique}
Ankner, Z.; Paul, M.; Cui, B.; Chang, J.~D.; and Ammanabrolu, P. 2024.
\newblock Critique-out-Loud Reward Models.
\newblock \emph{arXiv preprint arXiv:2408.11791}.

\bibitem[{Anthropic(2024)}]{anthropic2024claude}
Anthropic. 2024.
\newblock The claude 3 model family: Opus, sonnet, haiku.
\newblock \emph{Claude-3 Model Card}, 1: 1.

\bibitem[{Christiano et~al.(2017)Christiano, Leike, Brown, Martic, Legg, and Amodei}]{christiano2017deep}
Christiano, P.~F.; Leike, J.; Brown, T.; Martic, M.; Legg, S.; and Amodei, D. 2017.
\newblock Deep reinforcement learning from human preferences.
\newblock \emph{Advances in neural information processing systems}, 30.

\bibitem[{Cui et~al.(2024)Cui, Yuan, Ding, Yao, He, Zhu, Ni, Xie, Xie, Lin, Liu, and Sun}]{cui2024ultrafeedback}
Cui, G.; Yuan, L.; Ding, N.; Yao, G.; He, B.; Zhu, W.; Ni, Y.; Xie, G.; Xie, R.; Lin, Y.; Liu, Z.; and Sun, M. 2024.
\newblock {ULTRAFEEDBACK}: Boosting Language Models with Scaled {AI} Feedback.
\newblock In \emph{Proceedings of the 41st International Conference on Machine Learning}.

\bibitem[{DeepSeek-AI et~al.(2025)DeepSeek-AI, Guo, Yang, Zhang, Song, Zhang, Xu, Zhu, Ma, Wang, Bi, Zhang, Yu, Wu, Wu, Gou, Shao, Li, Gao, Liu, Xue, Wang, Wu, Feng, Lu, Zhao, Deng, Zhang, Ruan, Dai, Chen, Ji, Li, Lin, Dai, Luo, Hao, Chen, Li, Zhang, Bao, Xu, Wang, Ding, Xin, Gao, Qu, Li, Guo, Li, Wang, Chen, Yuan, Qiu, Li, Cai, Ni, Liang, Chen, Dong, Hu, Gao, Guan, Huang, Yu, Wang, Zhang, Zhao, Wang, Zhang, Xu, Xia, Zhang, Zhang, Tang, Li, Wang, Li, Tian, Huang, Zhang, Wang, Jin, Chen, Lu, Zhou, Chen, Ye, Wang, Yu, Zhou, Pan, Li, Zhou, Wu, Ye, Yun, Pei, Sun, Wang, Zeng, Zhao, Liu, Xiao, An, Liu, Wang, Chen, Nie, Cheng, Liu, Xie, Liu, Yang, Li, Su, Lin, Li, Jin, Shen, Chen, Sun, Wang, Song, Zhou, Wang, Shan, Li, Wang, Wei, Zhang, Xu, Li, Zhao, Sun, Wang, Yu, Zhang, Shi, Xiong, He, Piao, Wang, Tan, Ma, Liu, Guo, Ou, Wang, Gong, Zou, He, Xiong, Luo, You, Liu, Zhou, Zhu, Xu, Huang, Li, Zheng, Zhu, Ma, Tang, Zha, Yan, Ren, Ren, Sha, Fu, Xu, Xie, Zhang, Hao, Ma, Yan, Wu, Gu, Zhu, Liu, Li, Xie, Song, Pan, Huang,
  Xu, Zhang, and Zhang}]{guo2025deepseek}
DeepSeek-AI; Guo, D.; Yang, D.; Zhang, H.; Song, J.; Zhang, R.; Xu, R.; Zhu, Q.; Ma, S.; Wang, P.; Bi, X.; Zhang, X.; Yu, X.; Wu, Y.; Wu, Z.~F.; Gou, Z.; Shao, Z.; Li, Z.; Gao, Z.; Liu, A.; Xue, B.; Wang, B.; Wu, B.; Feng, B.; Lu, C.; Zhao, C.; Deng, C.; Zhang, C.; Ruan, C.; Dai, D.; Chen, D.; Ji, D.; Li, E.; Lin, F.; Dai, F.; Luo, F.; Hao, G.; Chen, G.; Li, G.; Zhang, H.; Bao, H.; Xu, H.; Wang, H.; Ding, H.; Xin, H.; Gao, H.; Qu, H.; Li, H.; Guo, J.; Li, J.; Wang, J.; Chen, J.; Yuan, J.; Qiu, J.; Li, J.; Cai, J.~L.; Ni, J.; Liang, J.; Chen, J.; Dong, K.; Hu, K.; Gao, K.; Guan, K.; Huang, K.; Yu, K.; Wang, L.; Zhang, L.; Zhao, L.; Wang, L.; Zhang, L.; Xu, L.; Xia, L.; Zhang, M.; Zhang, M.; Tang, M.; Li, M.; Wang, M.; Li, M.; Tian, N.; Huang, P.; Zhang, P.; Wang, Q.; Jin, R.~L.; Chen, R.; Lu, S.; Zhou, S.; Chen, S.; Ye, S.; Wang, S.; Yu, S.; Zhou, S.; Pan, S.; Li, S.~S.; Zhou, S.; Wu, S.; Ye, S.; Yun, T.; Pei, T.; Sun, T.; Wang, T.; Zeng, W.; Zhao, W.; Liu, W.; Xiao, W.~L.; An, W.; Liu, X.; Wang, X.; Chen, X.;
  Nie, X.; Cheng, X.; Liu, X.; Xie, X.; Liu, X.; Yang, X.; Li, X.; Su, X.; Lin, X.; Li, X.~Q.; Jin, X.; Shen, X.; Chen, X.; Sun, X.; Wang, X.; Song, X.; Zhou, X.; Wang, X.; Shan, X.; Li, Y.~K.; Wang, Y.~Q.; Wei, Y.~X.; Zhang, Y.; Xu, Y.; Li, Y.; Zhao, Y.; Sun, Y.; Wang, Y.; Yu, Y.; Zhang, Y.; Shi, Y.; Xiong, Y.; He, Y.; Piao, Y.; Wang, Y.; Tan, Y.; Ma, Y.; Liu, Y.; Guo, Y.; Ou, Y.; Wang, Y.; Gong, Y.; Zou, Y.; He, Y.; Xiong, Y.; Luo, Y.; You, Y.; Liu, Y.; Zhou, Y.; Zhu, Y.~X.; Xu, Y.; Huang, Y.; Li, Y.; Zheng, Y.; Zhu, Y.; Ma, Y.; Tang, Y.; Zha, Y.; Yan, Y.; Ren, Z.~Z.; Ren, Z.; Sha, Z.; Fu, Z.; Xu, Z.; Xie, Z.; Zhang, Z.; Hao, Z.; Ma, Z.; Yan, Z.; Wu, Z.; Gu, Z.; Zhu, Z.; Liu, Z.; Li, Z.; Xie, Z.; Song, Z.; Pan, Z.; Huang, Z.; Xu, Z.; Zhang, Z.; and Zhang, Z. 2025.
\newblock DeepSeek-R1: Incentivizing Reasoning Capability in LLMs via Reinforcement Learning.
\newblock arXiv:2501.12948.

\bibitem[{Fedus, Zoph, and Shazeer(2022)}]{fedus2022switch}
Fedus, W.; Zoph, B.; and Shazeer, N. 2022.
\newblock Switch Transformer: Scaling to Trillion Parameter Models with Simple and Efficient Sparsity.
\newblock \emph{Journal of Machine Learning Research}, 23(120): 1--39.

\bibitem[{Hendrycks et~al.(2021)Hendrycks, Burns, Kadavath, Arora, Basart, Tang, Song, and Steinhardt}]{hendrycks2021measuring}
Hendrycks, D.; Burns, C.; Kadavath, S.; Arora, A.; Basart, S.; Tang, E.; Song, D.; and Steinhardt, J. 2021.
\newblock Measuring Mathematical Problem Solving With the {MATH} Dataset.
\newblock In \emph{Thirty-fifth Conference on Neural Information Processing Systems Datasets and Benchmarks Track (Round 2)}.

\bibitem[{Lai et~al.(2024)Lai, Tian, Chen, Yang, Peng, and Jia}]{lai2024step}
Lai, X.; Tian, Z.; Chen, Y.; Yang, S.; Peng, X.; and Jia, J. 2024.
\newblock Step-dpo: Step-wise preference optimization for long-chain reasoning of llms.
\newblock \emph{arXiv preprint arXiv:2406.18629}.

\bibitem[{Lambert et~al.(2024)Lambert, Pyatkin, Morrison, Miranda, Lin, Chandu, Dziri, Kumar, Zick, Choi et~al.}]{lambert2024rewardbench}
Lambert, N.; Pyatkin, V.; Morrison, J.; Miranda, L.; Lin, B.~Y.; Chandu, K.; Dziri, N.; Kumar, S.; Zick, T.; Choi, Y.; et~al. 2024.
\newblock RewardBench: Evaluating Reward Models for Language Modeling.
\newblock \emph{arXiv preprint arXiv:2403.13787}.

\bibitem[{Liang et~al.(2019{\natexlab{a}})Liang, Zhang, Zhang, and Huang}]{liang2019asynchronous}
Liang, D.; Zhang, F.; Zhang, Q.; and Huang, X.-J. 2019{\natexlab{a}}.
\newblock Asynchronous deep interaction network for natural language inference.
\newblock In \emph{Proceedings of the 2019 Conference on Empirical Methods in Natural Language Processing and the 9th International Joint Conference on Natural Language Processing (EMNLP-IJCNLP)}, 2692--2700.

\bibitem[{Liang et~al.(2019{\natexlab{b}})Liang, Zhang, Zhang, Zhang, Fu, Peng, Gui, and Huang}]{liang2019adaptive}
Liang, D.; Zhang, F.; Zhang, W.; Zhang, Q.; Fu, J.; Peng, M.; Gui, T.; and Huang, X. 2019{\natexlab{b}}.
\newblock Adaptive multi-attention network incorporating answer information for duplicate question detection.
\newblock In \emph{Proceedings of the 42nd international ACM SIGIR conference on research and development in information retrieval}, 95--104.

\bibitem[{Lightman et~al.(2023)Lightman, Kosaraju, Burda, Edwards, Baker, Lee, Leike, Schulman, Sutskever, and Cobbe}]{lightman2023lets}
Lightman, H.; Kosaraju, V.; Burda, Y.; Edwards, H.; Baker, B.; Lee, T.; Leike, J.; Schulman, J.; Sutskever, I.; and Cobbe, K. 2023.
\newblock Let's Verify Step by Step.
\newblock \emph{arXiv preprint arXiv:2305.20050}.

\bibitem[{Liu et~al.(2024{\natexlab{a}})Liu, Zeng, Liu, Yan, He, Wang, Yan, Liu, and Zhou}]{liu2024skywork}
Liu, C.~Y.; Zeng, L.; Liu, J.; Yan, R.; He, J.; Wang, C.; Yan, S.; Liu, Y.; and Zhou, Y. 2024{\natexlab{a}}.
\newblock Skywork-reward: Bag of tricks for reward modeling in llms.
\newblock \emph{arXiv preprint arXiv:2410.18451}.

\bibitem[{Liu et~al.(2025{\natexlab{a}})Liu, Liang, Shan, Liu, Liu, Wu, Li, Wu, Miao, Shen et~al.}]{liu2025structural}
Liu, X.; Liang, D.; Shan, H.; Liu, P.; Liu, Y.; Wu, M.; Li, Y.; Wu, X.; Miao, L.; Shen, J.; et~al. 2025{\natexlab{a}}.
\newblock Structural Reward Model: Enhancing Interpretability, Efficiency, and Scalability in Reward Modeling.
\newblock In \emph{Proceedings of the 2025 Conference on Empirical Methods in Natural Language Processing: Industry Track}, 672--685.

\bibitem[{Liu et~al.(2024{\natexlab{b}})Liu, Yao, Min, Cao, Hou, and Li}]{liu2024rmbenchbenchmarkingrewardmodels}
Liu, Y.; Yao, Z.; Min, R.; Cao, Y.; Hou, L.; and Li, J. 2024{\natexlab{b}}.
\newblock RM-Bench: Benchmarking Reward Models of Language Models with Subtlety and Style.
\newblock arXiv:2410.16184.

\bibitem[{Liu et~al.(2025{\natexlab{b}})Liu, Wang, Xu, Ma, Ruan, Li, Liu, and Wu}]{liu2025inference}
Liu, Z.; Wang, P.; Xu, R.; Ma, S.; Ruan, C.; Li, P.; Liu, Y.; and Wu, Y. 2025{\natexlab{b}}.
\newblock Inference-Time Scaling for Generalist Reward Modeling.
\newblock \emph{arXiv preprint arXiv:2504.02495}.

\bibitem[{Luo et~al.(2024)}]{luo2024improve}
Luo, L.; et~al. 2024.
\newblock Improve Mathematical Reasoning in Language Models by Automated Process Supervision.
\newblock \emph{arXiv preprint arXiv:2406.06592}.

\bibitem[{McAleese et~al.(2024)McAleese, Pokorny, Uribe, Nitishinskaya, Trebacz, and Leike}]{mcaleese2024llm}
McAleese, N.; Pokorny, R.~M.; Uribe, J. F.~C.; Nitishinskaya, E.; Trebacz, M.; and Leike, J. 2024.
\newblock LLM Critics Help Catch LLM Bugs.
\newblock \emph{arXiv preprint arXiv:2407.00215}.

\bibitem[{OpenAI et~al.(2024)OpenAI, :, Hurst, Lerer, Goucher, Perelman, Ramesh, Clark, Ostrow, Welihinda, Hayes, Radford, Mądry, Baker-Whitcomb, Beutel, Borzunov, Carney, Chow, Kirillov, Nichol, Paino, Renzin, Passos, Kirillov, Christakis, Conneau, Kamali, Jabri, Moyer, Tam, Crookes, Tootoochian, Tootoonchian, Kumar, Hallacy, Koch, Gibson, Kim, Choi, McLeavey, Hesse, Fischer, Winter, Czarnecki, Jarvis, Wei, Koumouzelis, Sherburn, Kappler, Levin, Levy, Carr, Farhi, Mely, Robinson, Sasaki, Jin, Valladares, Tsipras, Li, Nguyen, Findlay, Oiwoh, Wong, Asdar, Proehl, Yang, Puckett, Nachum, Okelola, Boiko, Murk, Jaffe, Watkins, Godement, Campbell-Moore, Chao, McMillan, Belov, Su, Bak, Bakkum, Deng, Dolan, Hoeschele, Welinder, Tillet, Pronin, Tillet, Dhariwal, Yuan, Dias, Lim, Arora, Troll, Lin, Lopes, Puri, Miyara, Leike, Gaubert, Zamani, Wang, Donnelly, Honsby, Smith, Sahai, Phene, Papay, Narayanan, Coffey, Lee, Hall, Balaji, Broda, Stramer, Xu, Gogineni, Christianson, Sanders, Patwardhan, Cunninghman, Degry,
  Dimson, Raoux, Shadwell, Zheng, Underwood, Markov, Sherbakov, Rubin, Stasi, Kaftan, Heywood, Peterson, Walters, Eloundou, Qi, Moeller, Monaco, Kuo, Fomenko, Chang, Zheng, Zhou, Manassra, Sheu, Zaremba, Patil, Qian, Kim, Cheng, Zhang, He, Zhang, Jin, Dai, and Malkov}]{openai2024gpt4technicalreport}
OpenAI; :; Hurst, A.; Lerer, A.; Goucher, A.~P.; Perelman, A.; Ramesh, A.; Clark, A.; Ostrow, A.; Welihinda, A.; Hayes, A.; Radford, A.; Mądry, A.; Baker-Whitcomb, A.; Beutel, A.; Borzunov, A.; Carney, A.; Chow, A.; Kirillov, A.; Nichol, A.; Paino, A.; Renzin, A.; Passos, A.~T.; Kirillov, A.; Christakis, A.; Conneau, A.; Kamali, A.; Jabri, A.; Moyer, A.; Tam, A.; Crookes, A.; Tootoochian, A.; Tootoonchian, A.; Kumar, A.; Hallacy, C.; Koch, C.; Gibson, C.; Kim, C.; Choi, C.; McLeavey, C.; Hesse, C.; Fischer, C.; Winter, C.; Czarnecki, C.; Jarvis, C.; Wei, C.; Koumouzelis, C.; Sherburn, D.; Kappler, D.; Levin, D.; Levy, D.; Carr, D.; Farhi, D.; Mely, D.; Robinson, D.; Sasaki, D.; Jin, D.; Valladares, D.; Tsipras, D.; Li, D.; Nguyen, D.~P.; Findlay, D.; Oiwoh, E.; Wong, E.; Asdar, E.; Proehl, E.; Yang, E.; Puckett, N.; Nachum, O.; Okelola, O.; Boiko, O.; Murk, O.; Jaffe, O.; Watkins, O.; Godement, O.; Campbell-Moore, O.; Chao, P.; McMillan, P.; Belov, P.; Su, P.; Bak, P.; Bakkum, P.; Deng, P.; Dolan, P.;
  Hoeschele, P.; Welinder, P.; Tillet, P.; Pronin, P.; Tillet, P.; Dhariwal, P.; Yuan, Q.; Dias, R.; Lim, R.; Arora, R.; Troll, R.; Lin, R.; Lopes, R.~G.; Puri, R.; Miyara, R.; Leike, R.; Gaubert, R.; Zamani, R.; Wang, R.; Donnelly, R.; Honsby, R.; Smith, R.; Sahai, R.; Phene, S.; Papay, S.; Narayanan, S.; Coffey, S.; Lee, S.; Hall, S.; Balaji, S.; Broda, T.; Stramer, T.; Xu, T.; Gogineni, T.; Christianson, T.; Sanders, T.; Patwardhan, T.; Cunninghman, T.; Degry, T.; Dimson, T.; Raoux, T.; Shadwell, T.; Zheng, T.; Underwood, T.; Markov, T.; Sherbakov, T.; Rubin, T.; Stasi, T.; Kaftan, T.; Heywood, T.; Peterson, T.; Walters, T.; Eloundou, T.; Qi, V.; Moeller, V.; Monaco, V.; Kuo, V.; Fomenko, V.; Chang, W.; Zheng, W.; Zhou, W.; Manassra, W.; Sheu, W.; Zaremba, W.; Patil, Y.; Qian, Y.; Kim, Y.; Cheng, Y.; Zhang, Y.; He, Y.; Zhang, Y.; Jin, Y.; Dai, Y.; and Malkov, Y. 2024.
\newblock GPT-4o System Card.
\newblock arXiv:2410.21276.

\bibitem[{Ouyang et~al.(2022)Ouyang, Wu, Jiang, Almeida, Wainwright, Mishkin, Zhang, Agarwal, Slama, Ray et~al.}]{ouyang2022training}
Ouyang, L.; Wu, J.; Jiang, X.; Almeida, D.; Wainwright, C.; Mishkin, P.; Zhang, C.; Agarwal, S.; Slama, K.; Ray, A.; et~al. 2022.
\newblock Training language models to follow instructions with human feedback.
\newblock \emph{Advances in neural information processing systems}, 35: 27730--27744.

\bibitem[{Park et~al.(2024)Park, Jwa, Meiying, Kim, and Choi}]{park2024offsetbias}
Park, J.; Jwa, S.; Meiying, R.; Kim, D.; and Choi, S. 2024.
\newblock {O}ffset{B}ias: Leveraging Debiased Data for Tuning Evaluators.
\newblock In \emph{Findings of the Association for Computational Linguistics: EMNLP 2024}.

\bibitem[{Rafailov et~al.(2024)Rafailov, Sharma, Mitchell, Ermon, Manning, and Finn}]{rafailov2024directpreferenceoptimizationlanguage}
Rafailov, R.; Sharma, A.; Mitchell, E.; Ermon, S.; Manning, C.~D.; and Finn, C. 2024.
\newblock Direct Preference Optimization: Your Language Model is Secretly a Reward Model.
\newblock arXiv:2305.18290.

\bibitem[{Shao et~al.(2024)Shao, Wang, Feng, Zhu, Gan, Wang et~al.}]{shao2024deepseek}
Shao, Z.; Wang, P.; Feng, Q.; Zhu, H.; Gan, Z.; Wang, S.; et~al. 2024.
\newblock {DeepSeekMath: Pushing the Limits of Mathematical Reasoning in Open Language Models}.
\newblock \emph{arXiv preprint arXiv:2402.03300}, 1(1): 1--35.

\bibitem[{Stiennon et~al.(2020)Stiennon, Ouyang, Wu, Ziegler, Lowe, Voss, Radford, Amodei, and Christiano}]{stiennon2020learning}
Stiennon, N.; Ouyang, L.; Wu, J.; Ziegler, D.; Lowe, R.; Voss, C.; Radford, A.; Amodei, D.; and Christiano, P.~F. 2020.
\newblock Learning to summarize with human feedback.
\newblock 33: 3008--3021.

\bibitem[{Wang et~al.(2024{\natexlab{a}})Wang, Xiong, Xie, Zhao, and Zhang}]{wang2024interpretable}
Wang, H.; Xiong, W.; Xie, T.; Zhao, H.; and Zhang, T. 2024{\natexlab{a}}.
\newblock Interpretable Preferences via Multi-Objective Reward Modeling and Mixture-of-Experts.
\newblock In \emph{Findings of the Association for Computational Linguistics: EMNLP 2024}.

\bibitem[{Wang et~al.(2024{\natexlab{b}})Wang, Li, Shao, Xu, Dai, Li, Chen, Wu, and Sui}]{wang2024math}
Wang, P.; Li, L.; Shao, Z.; Xu, R.; Dai, D.; Li, Y.; Chen, D.; Wu, Y.; and Sui, Z. 2024{\natexlab{b}}.
\newblock Math-Shepherd: Verify and Reinforce LLMs Step-by-step without Human Annotations.
\newblock \emph{Proceedings of the 62nd Annual Meeting of the Association for Computational Linguistics}.

\bibitem[{Wang et~al.(2022)Wang, Liang, Song, Li, and Wu}]{wang2022dabert}
Wang, S.; Liang, D.; Song, J.; Li, Y.; and Wu, W. 2022.
\newblock Dabert: Dual attention enhanced bert for semantic matching.
\newblock \emph{arXiv preprint arXiv:2210.03454}.

\bibitem[{Wang et~al.(2024{\natexlab{c}})Wang, Kulikov, Golovneva, Yu, Yuan, Dwivedi-Yu, Pang, Fazel-Zarandi, Weston, and Li}]{wang2024self}
Wang, T.; Kulikov, I.; Golovneva, O.; Yu, P.; Yuan, W.; Dwivedi-Yu, J.; Pang, R.~Y.; Fazel-Zarandi, M.; Weston, J.; and Li, X. 2024{\natexlab{c}}.
\newblock Self-taught evaluators.
\newblock \emph{arXiv preprint arXiv:2408.02666}.

\bibitem[{Wang et~al.(2023)Wang, Wei, Schuurmans, Le, Chi, Narang, Chowdhery, and Zhou}]{wang2023selfconsistencyimproveschainthought}
Wang, X.; Wei, J.; Schuurmans, D.; Le, Q.; Chi, E.; Narang, S.; Chowdhery, A.; and Zhou, D. 2023.
\newblock Self-Consistency Improves Chain of Thought Reasoning in Language Models.
\newblock arXiv:2203.11171.

\bibitem[{Wang, Liang, and Peng(2025)}]{wang2025not}
Wang, Y.; Liang, D.; and Peng, M. 2025.
\newblock Not all parameters are created equal: Smart isolation boosts fine-tuning performance.
\newblock \emph{arXiv preprint arXiv:2508.21741}.

\bibitem[{Wang et~al.(2024{\natexlab{d}})Wang, Bukharin, Delalleau, Egert, Shen, Zeng, Kuchaiev, and Dong}]{wang2024helpsteer2preference}
Wang, Z.; Bukharin, A.; Delalleau, O.; Egert, D.; Shen, G.; Zeng, J.; Kuchaiev, O.; and Dong, Y. 2024{\natexlab{d}}.
\newblock HelpSteer2-Preference: Complementing Ratings with Preferences.
\newblock arXiv:2410.01257.

\bibitem[{Wei et~al.(2022)Wei, Wang, Schuurmans, Bosma, Ichter, Xia, Chi, Le, and Zhou}]{wei2022chain}
Wei, J.; Wang, X.; Schuurmans, D.; Bosma, M.; Ichter, B.; Xia, F.; Chi, E.~H.; Le, Q.~V.; and Zhou, D. 2022.
\newblock Chain of Thought Prompting Elicits Reasoning in Large Language Models.
\newblock In \emph{Advances in Neural Information Processing Systems}.

\bibitem[{Wu et~al.(2025)Wu, Sun, Li, Welleck, and Yang}]{wu2025inferencescalinglawsempirical}
Wu, Y.; Sun, Z.; Li, S.; Welleck, S.; and Yang, Y. 2025.
\newblock Inference Scaling Laws: An Empirical Analysis of Compute-Optimal Inference for Problem-Solving with Language Models.
\newblock arXiv:2408.00724.

\bibitem[{Yao et~al.(2024)Yao, Yu, Zhao, Shafran, Griffiths, Cao, and Narasimhan}]{yao2024tree}
Yao, S.; Yu, D.; Zhao, J.; Shafran, I.; Griffiths, T.~L.; Cao, Y.; and Narasimhan, K. 2024.
\newblock Tree of Thoughts: Deliberate Problem Solving with Large Language Models.
\newblock \emph{Advances in Neural Information Processing Systems}, 36.

\bibitem[{Yu et~al.(2024)Yu, Chen, Zhang, Tan, Zhu, Pang, Qian, Wang, Gururangan, Zhang et~al.}]{yu2024self}
Yu, Y.; Chen, Z.; Zhang, A.; Tan, L.; Zhu, C.; Pang, R.~Y.; Qian, Y.; Wang, X.; Gururangan, S.; Zhang, C.; et~al. 2024.
\newblock Self-Generated Critiques Boost Reward Modeling for Language Models.
\newblock arXiv:2411.16646.

\bibitem[{Zhang et~al.(2022)Zhang, Zhang, Li, and Smola}]{zhang2022automaticchainthoughtprompting}
Zhang, Z.; Zhang, A.; Li, M.; and Smola, A. 2022.
\newblock Automatic Chain of Thought Prompting in Large Language Models.
\newblock arXiv:2210.03493.

\bibitem[{Zhang et~al.(2025)Zhang, Zheng, Wu, Zhang, Lin, Yu, Liu, Zhou, and Lin}]{zhang2025lessonsdevelopingprocessreward}
Zhang, Z.; Zheng, C.; Wu, Y.; Zhang, B.; Lin, R.; Yu, B.; Liu, D.; Zhou, J.; and Lin, J. 2025.
\newblock The Lessons of Developing Process Reward Models in Mathematical Reasoning.
\newblock arXiv:2501.07301.

\bibitem[{Zheng et~al.(2024)Zheng, Zhang, Zhang, Lin, Lu, Yu, Liu, Zhou, and Lin}]{zheng2024processbench}
Zheng, C.; Zhang, Z.; Zhang, B.; Lin, R.; Lu, K.; Yu, B.; Liu, D.; Zhou, J.; and Lin, J. 2024.
\newblock ProcessBench: Identifying Process Errors in Mathematical Reasoning.
\newblock \emph{arXiv preprint arXiv:2412.06559}.

\bibitem[{Zhou et~al.(2023)Zhou, Schärli, Hou, Wei, Scales, Wang, Schuurmans, Cui, Bousquet, Le, and Chi}]{zhou2023leasttomostpromptingenablescomplex}
Zhou, D.; Schärli, N.; Hou, L.; Wei, J.; Scales, N.; Wang, X.; Schuurmans, D.; Cui, C.; Bousquet, O.; Le, Q.; and Chi, E. 2023.
\newblock Least-to-Most Prompting Enables Complex Reasoning in Large Language Models.
\newblock arXiv:2205.10625.

\bibitem[{Zhou et~al.(2025)Zhou, Zheng, Wang, Xi, Dou, Bao, Shen, Xiong, Fan, Mou, Zheng, Gui, Zhang, and Huang}]{zhou2025rmbcomprehensivelybenchmarkingreward}
Zhou, E.; Zheng, G.; Wang, B.; Xi, Z.; Dou, S.; Bao, R.; Shen, W.; Xiong, L.; Fan, J.; Mou, Y.; Zheng, R.; Gui, T.; Zhang, Q.; and Huang, X. 2025.
\newblock RMB: Comprehensively Benchmarking Reward Models in LLM Alignment.

\bibitem[{Zhu et~al.(2023)Zhu, Frick, Wu, Zhu, and Jiao}]{starling2023}
Zhu, B.; Frick, E.; Wu, T.; Zhu, H.; and Jiao, J. 2023.
\newblock Starling-7B: Improving LLM Helpfulness \& Harmlessness with RLAIF.

\bibitem[{Ziegler et~al.(2019)Ziegler, Stiennon, Wu, Brown, Radford, Amodei, Christiano, and Irving}]{ziegler2019fine}
Ziegler, D.~M.; Stiennon, N.; Wu, J.; Brown, T.~B.; Radford, A.; Amodei, D.; Christiano, P.; and Irving, G. 2019.
\newblock Fine-tuning language models from human preferences.
\newblock \emph{arXiv preprint arXiv:1909.08593}.

\end{thebibliography}
\end{document}